\documentclass[%
 aip,
 jmp,%
 amsmath,amssymb,
 reprint,%
]{revtex4-1}

\usepackage{graphicx}
\usepackage{dcolumn}
\usepackage{bm}
\usepackage{subfigure}

\usepackage{soul}

\usepackage{amsmath,amssymb,amsfonts}
\DeclareMathOperator*{\argmax}{arg\,max}

\usepackage{graphics}
\usepackage{color}
\usepackage{xcolor}
\definecolor{rev1}{rgb}{0,0,0}

\usepackage{algorithm}
\usepackage{algpseudocode}

\algnewcommand\server{\item[\textbf{Server execution:}]}%
\algnewcommand\client{\item[\textbf{ClientUpdate($k,w$):}]}%

\usepackage{enumitem}
\setlist[itemize]{leftmargin=*}
\setlist[enumerate]{leftmargin=*}

\usepackage{comment}
\usepackage{hyperref}
\usepackage{lipsum}
\usepackage{tabularx}
\usepackage{multirow}
\usepackage{mathrsfs}
\usepackage{stackengine}

\usepackage{accents}

\begin{document}






\title{Prospects of federated machine learning in fluid dynamics}




\author{Omer San}%
\email{osan@okstate.edu}
\author{Suraj Pawar}
\affiliation{ 
School of Mechanical \& Aerospace Engineering, Oklahoma State University, Stillwater, OK 74078, USA.
}%


\author{Adil Rasheed}%
\affiliation{ 
Department of Engineering Cybernetics, Norwegian University of Science and Technology, N-7465, Trondheim, Norway.
 }%




\date{\today}

\begin{abstract}
Physics-based models have been mainstream in fluid dynamics for developing predictive models. In recent years, machine learning has offered a renaissance to the fluid community due to the rapid developments in data science, processing units, neural network based technologies, and sensor adaptations. So far in many applications in fluid dynamics, machine learning approaches have been mostly focused on a standard process that requires centralizing the training data on a designated machine or in a data center. In this letter, we present a federated machine learning approach that enables localized clients to collaboratively learn an aggregated and shared predictive model while keeping all the training data on each edge device. We demonstrate the feasibility and prospects of such decentralized learning approach with an effort to forge a deep learning surrogate model for reconstructing spatiotemporal fields. Our results indicate that federated machine learning might be a viable tool for designing highly accurate predictive decentralized digital twins relevant to fluid dynamics.
 
\end{abstract}


\maketitle

In many complex systems involving fluid flows, computing a physics-based model might be prohibitive, especially when our simulations are compatible with the timescales of natural phenomena. Consequently, there is an ever-growing interest in generating surrogate or reduced order models \cite{ahmed2021closures}. It has also been envisioned that a digital twin capable of accurately representing the physical system could offer a better value proposition to specific applications and stakeholders \cite{rasheed2020digital}. The role of this digital twin might be to provide descriptive, diagnostic, predictive, or prescriptive guidelines for a better-informed decision. \textcolor{rev1}{The \textit{market pull} created by digital twin-like technologies coupled with the \textit{technology push}} provided by significant advances in machine learning (ML) and artificial intelligence (AI), advanced and cost-effective sensor technologies, readily available computational resources, and opensource ML libraries have accelerated ML penetration in domain sciences like never before. The last decade has seen an exponential growth of data-driven modeling technologies (e.g., deep neural networks) that might be key enablers for improving the modeling accuracy of geophysical fluid systems \citep{san2021hybrid}. A recent workshop held by NASA Advanced Information Systems Technology Program and Earth Science Information Partners on ML adoption\citep{nasa2018} identified the following guidelines, among many others, in this area:
\begin{itemize}
    \item Cutting edge ML algorithms and techniques need to be available, packaged in some way and well understood so as to be usable.
    \item Computer security implementations are outdated and uncooperative with science investigations. Research in making computational resources secure and yet easily usable would be valuable.
\end{itemize}

\textcolor{rev1}{One of the fluid flow problems that ML and AI can positively impact is weather forecasting. Big data will be the key to making the digital twins of the natural environments a reality.} In addition to the data from forecasting models and dedicated weather stations, it can be expected that there will be an unprecedented penetration of smart devices (e.g., smart weather stations, smartphones, and smartwatches), and contributions from crowdsourcing. For example, by 2025, there will be more than 7 billion smartphones worldwide. This number is much more significant than the paltry (over 10,000) official meteorological stations around the world \cite{mildrexler2011global}. While analyzing and utilizing data only from a few edge devices might not yield accurate predictions, processing data from many smart and connected devices equipped with sensors might be a game changer in weather monitoring and prediction. In their recent report, \citet{o2021service} highlighted that the Weather Company utilizes data from over 250,000 personal weather stations.
Moreover, \citet{chapman2017can} discussed how the crowdsourcing data-driven modeling paradigm could take meteorological science to a new level using smart Netatmo weather stations. As more attention shifts to smart and connected \emph{internet of things} devices, security and privacy implications of such smart weather stations have also been discussed \cite{sivaraman2018smart}. \textcolor{rev1}{Additionally, big data will come with its own challenges characterized by 10 Vs \cite{fote2020big}. The 10 Vs imply large volume, velocity, variety, veracity, value, validity, variability, venue, vocabulary, and vagueness. Volume refers to the size of data, velocity refers to the data generation rate, variety refers to the data type, veracity refers to the data quality and accuracy, value refers to the data usefulness, validity refers to the data quality and governance, variability refers to the dynamic, evolving behavior in the data source, venue refers to the heterogeneous data from multiple sources, and vocabulary refers to the semantics describing data structure. Finally, vagueness refers to the confusion over the meaning of data and tools used. In the weather forecast and many other processes, we foresee that all these problems will have to be addressed.}

To this end, in this letter, we focus on the statistical learning part and introduce a distributed training approach to generate autoencoder models that are relevant to the nonlinear dimensionality reduction of spatiotemporally distributed data sets. We aim at exploring the feasibility of such a decentralized learning framework to model complex spatiotemporal systems in which local data samples are held in edge devices. The case handled here is relatively simple but that was completely intentional as it eases the communication and dissemination of the work to a larger audience. Specifically, we put forth a federated ML framework considering the Kuramoto–Sivashinsky (KS) system\cite{armbruster1989kuramoto,holmes2012turbulence}, which is known for its irregular or chaotic behavior. 

\begin{figure}[ht]
\centering
\includegraphics[width=1.0\linewidth]{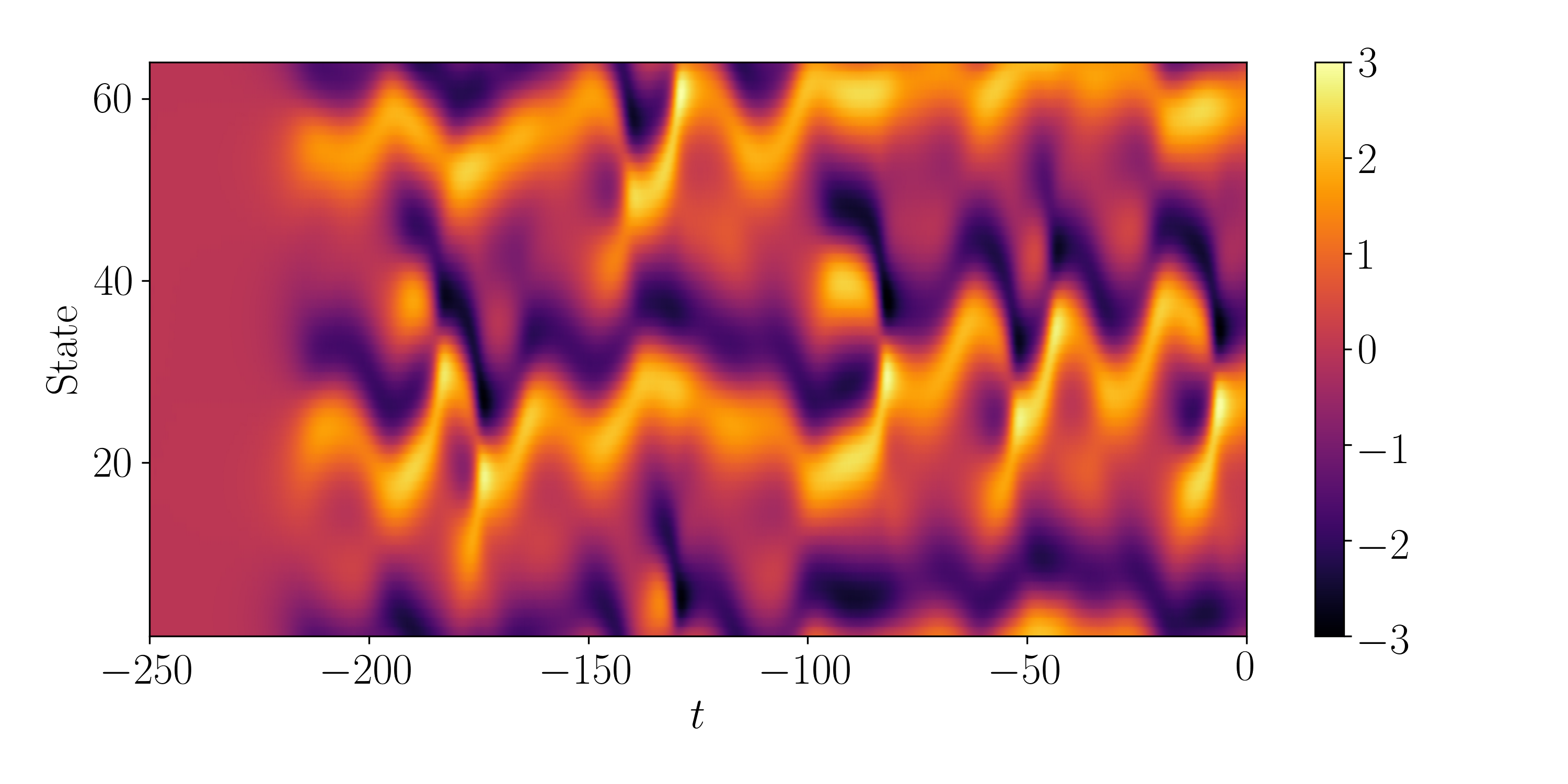}
\vspace{-0.3in}
\caption{The evolution of the KS system illustrating the spatiotemporal field data at the initial transient period.}
\vspace{-0.1in}
\label{fig:initial}
\end{figure}

This system has been derived to describe diffusion-induced chaotic behavior in reaction systems \cite{kuramoto1978diffusion}, hydrodynamic instabilities in laminar flames \cite{sivashinsky1977nonlinear}, \textcolor{rev1}{phase dynamics of nonlinear Alfv{\'e}n waves\cite{rempel2004intermittent}} as well as nonlinear saturation of fluctuation potentials in plasma physics \cite{laquey1975nonlinear}. Due to its systematic route to chaos, the KS system has attracted much attention recently to test the feasibility of emerging ML approaches specifically designed to capture complex spatiotemporal dynamics (see, for example, \citet{gonzalez1998identification,pathak2018model,vlachas2018data,linot2020deep,vlachas2022multiscale}). The KS equation with $L$-periodic boundary conditions can be written as 
\begin{equation}
    \frac{\partial u}{\partial t} = -\frac{\partial^4  u}{\partial x^4} - \frac{\partial^2 u}{\partial x^2} - u \frac{\partial u}{\partial x},
    \label{eq:ks}
\end{equation}
on a spatial domain $x \in [0, L]$, where the dynamics undergo a hierarchy of bifurcations as the spatial domain size $L$ is increased, building up the chaotic behavior. Here, we perform the underlying numerical experiments with $L=22$ to generate our spatiotemporal data set. Equation~\ref{eq:ks} is solved using the fourth-order method for stiff partial differential equations \cite{kassam2005fourth} with the spatial grid size of $N=64$. The random initial condition is assigned at time $t=-250$ and the solution is evolved with a time step of $2.5 \times 10^{-3}$ up to $t=0$. The trajectory of the KS system in the initial transient period is shown in Figure~\ref{fig:initial}. Using the solution at time $t=0$ as the initial condition, the KS system is evolved till $t=2500$. The data is sampled at a time step of 0.25 and these 10,000 samples are used for training and validation. For the testing purpose, the data from $t=2500$ to $t=3750$ is utilized.

In this work, the federated ML is demonstrated for an autoencoder which is a powerful approach for obtaining the latent space on a nonlinear manifold. The autoencoder is composed of the encoder and a decoder, where the encoder maps an input to a low-dimensional latent space and the decoder performs the inverse mapping from latent space variables to the original dimension at the output. If we denote the encoder function as $\eta(w)$ and a decoder function is defined as $\xi(w)$, then we can represent the manifold learning as follows
\begin{align}
    \eta, \xi &= \argmax_{\eta, \xi} ~\lVert \boldsymbol{u} - (\eta \circ \xi)\boldsymbol{u} \rVert, \\
    \eta &: \boldsymbol{u} \in \mathbb{R}^N \rightarrow \boldsymbol{z} \in \mathbb{R}^R, \\
    \xi &: \boldsymbol{z} \in \mathbb{R}^R \rightarrow \boldsymbol{u} \in \mathbb{R}^N,
\end{align}
where $\boldsymbol{z}$ represent the low dimensional latent space and $R$ is the dimensionality of the latent space. 

\begin{figure*}[ht]
\centering
\includegraphics[width=0.80\linewidth]{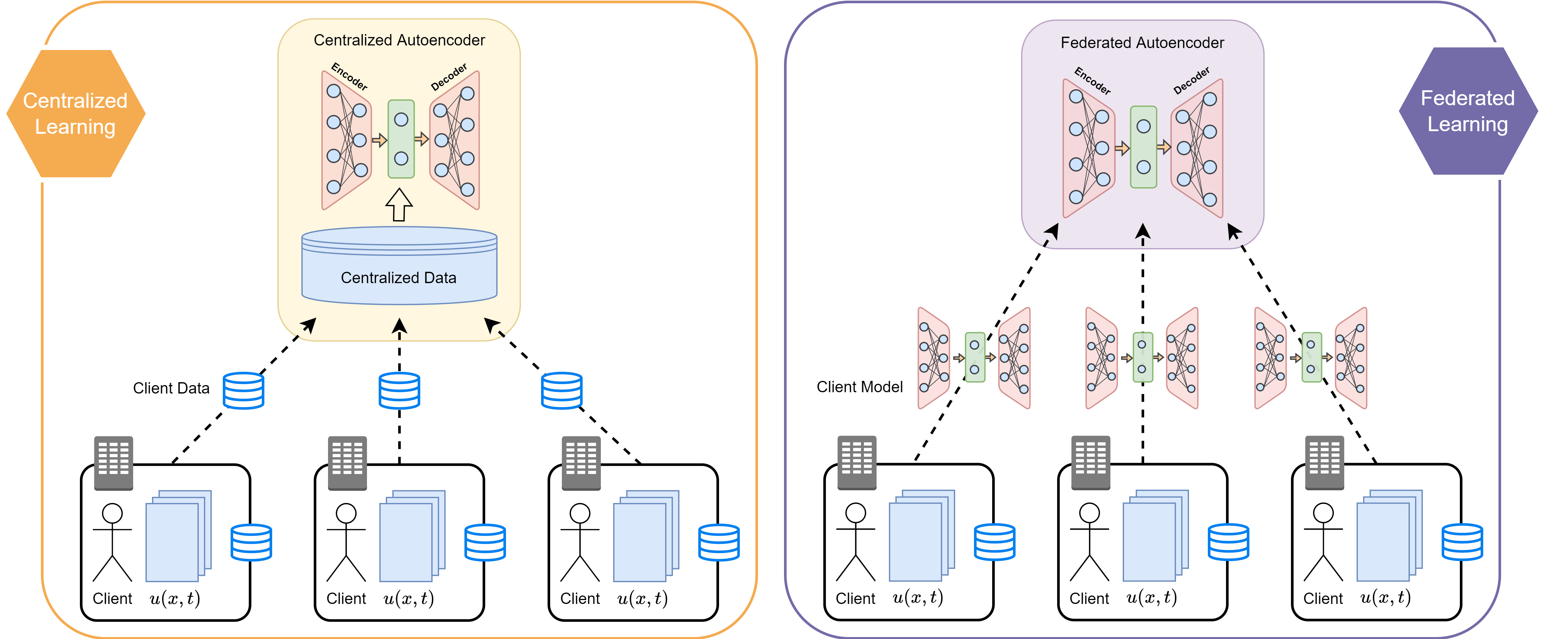}
\caption{Overview and schematic illustrations of the centralized and federated ML approaches.} \vspace{-0.1in}
\label{fig:fml}
\end{figure*}

We closely follow the seminal work in federated learning \cite{mcmahan2017communication}, which introduces a federated averaging algorithm where clients collaboratively train a shared model. Figure~\ref{fig:fml} contrasts the federated learning approach with the centralized method. In the centralized method, the local dataset is transferred from clients to a central server and the model is trained using centrally stored data. In case of the federated learning, the local dataset is never transferred from clients to a server. Instead, each client computes an update to the global model maintained by the server based on the local dataset, and only this update to the model is communicated. The federated averaging algorithm assumes that there is a fixed set of $K$ clients with a fixed local dataset and a synchronous update scheme is applied in rounds of communications. At the beginning of each communication round, the central server sends the global state of the model (i.e., the current model parameters) to each of the clients. Each client computes the update to the global model based on the global state and local dataset and this update is sent to a server. The server then updates the global state of the model based on the local updates received from all clients, and this process continues. The objective function for a federated averaging algorithm can be written as follows 
\begin{equation}
    f(w) = \sum_{k=1}^{K} \frac{n_k}{n}F_k(w) \quad \text{where} \quad F_k(w)=\frac{1}{n_k} \sum_{i \in \mathcal{P}_k} f_i(w),
\end{equation}
$\mathcal{P}_k$ is the data on the $k$th client, \textcolor{rev1}{$n_k$ is the cardinality of $\mathcal{P}_k$}, and $f_i(w)=l(x_i,y_i;w)$ is the loss of the prediction on example $(x_i,y_i)$. The above aggregation protocol can be applied to any ML algorithm. In this work, we use the autoencoder for nonlinear dimensionality reduction\cite{ahmed2021nonlinear}, and the complete pseudo-code for deep learning models in a federated setting is provided in Algorithm~\ref{alg:fl}. We highlight that the approach we utilize in our study simply weights edge devices proportionally by the data they own. More advanced approaches can be considered to mitigate such limitations \cite{li2020federated,li2019fair,fallah2020personalized,deng2020adaptive,tan2022towards}, but that is beyond the scope of this letter. 

\begin{figure}[ht]
\centering
\includegraphics[width=1.0\linewidth]{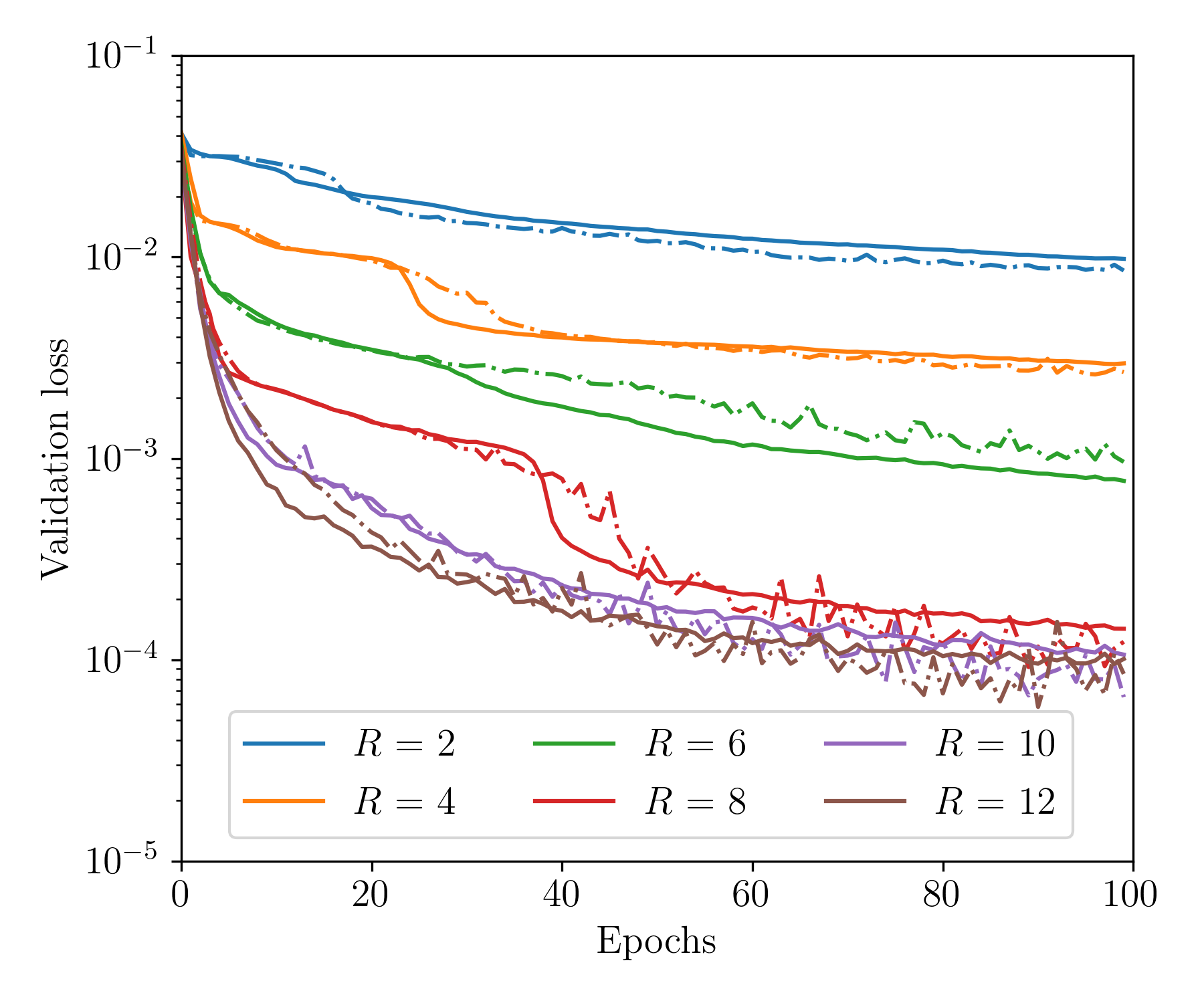}
\vspace{-0.3in}
\caption{Validation loss during training. Here, dashed line corresponds to centralized learning and solid lines are for federated learning.}
\label{fig:loss}
\vspace{-0.1in}
\end{figure}

\begin{figure}[ht]
\centering
\includegraphics[width=1.0\linewidth]{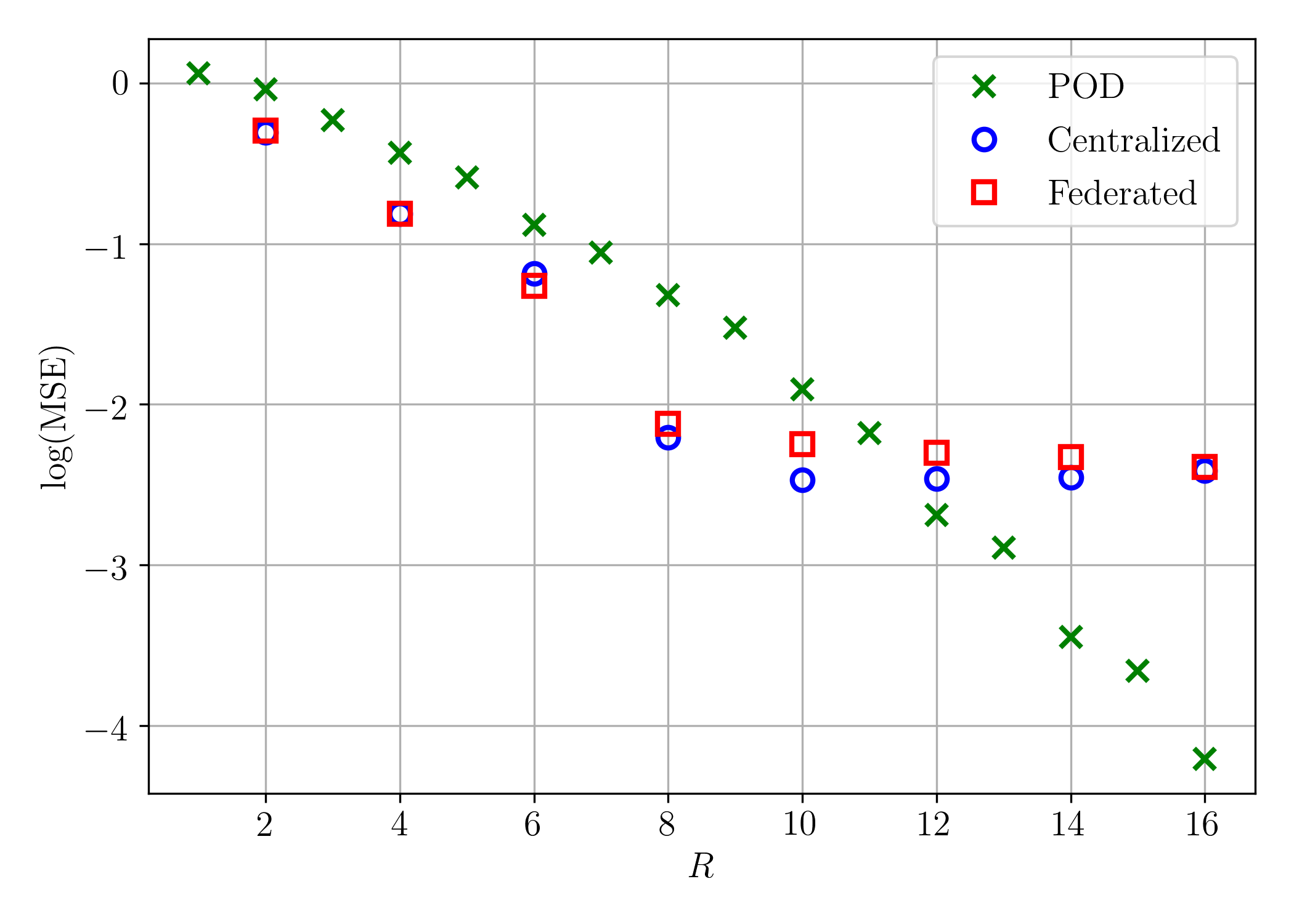}
\vspace{-0.3in}
\caption{Reconstruction mean squared error (MSE) on the test data.}
\label{fig:mse}
\vspace{-0.1in}
\end{figure}

Following the work of \citet{vlachas2022multiscale}, we first validate the centralized approach by varying $R$. For the federated learning, we use $K=10$ clients, and each client model is trained for $E=1$ local epoch with a batch size $B=32$. For a fair comparison, the batch size of 320 is utilized for training the centralized autoencoder. The validation loss for the centralized and federated autoencoder with different dimensionality of the latent space is depicted in Figure~\ref{fig:loss} and we see that both the losses converge to very similar values. This shows that there is no significant loss in accuracy due to federated learning compared to centralized learning. As shown in Figure~\ref{fig:mse}, the reconstruction error for both centralized and federated autoencoders saturates around $R=8$ modes. Figure~\ref{fig:mse} also demonstrates that a linear approach based on the proper orthogonal decomposition (POD) (see, e.g., \citet{ahmed2021closures,ahmed2019memory,pawar2021model,pawar2020data,san2014proper,san2015stabilized}) requires significantly more modes to represent underlying flow dynamics with the same accuracy. Our observations, which are consistent with previous works \cite{vlachas2022multiscale,linot2020deep,cvitanovic2010state,robinson1994inertial}, suggest that the latent space dynamics lies effectively on a manifold with $R=8$ dimensions. \textcolor{rev1}{Although our analysis includes a global POD approach for comparison purpose, we may consider to apply a localized POD approach\cite{tadmor2008fast,san2015principal,ahmed2018stabilized} for improved modal representation. Instead of a detailed POD analysis here, our work rather aims primarily at demonstrating the potential of federated learning in fluid mechanics as opposed to centralized learning.}

\begin{algorithm}[H]
\caption{Federated averaging algorithm. $B$ is the local minibatch size, $E$ is the number of local epochs, and $\alpha$ is the learning rate. }\label{alg:fl}
\begin{algorithmic}[]
\server{}
\State initialize $w_0$
\For{$t=1,2,\dots$}
\For{each client k}
\State{$w_{t+1}^k \leftarrow$ ClientUpdate($k,w_t$)} 
\EndFor
\State{$w_{t+1} \leftarrow \sum_{k=1}^K\frac{n_k}{n} w_{t+1}^k$}
\EndFor
\vspace{10pt}
\client{}
\State{$\mathcal{B} \leftarrow $ (split $\mathcal{P}_k$ into batches of size B)}
\For{each local epoch $i$ from 1 to $E$}
\For{batch $b \in \mathcal{B}$}
\State{$w \leftarrow w - \alpha \nabla l(w;b)$}
\EndFor
\EndFor
\State{return $w$ to a server}
\end{algorithmic}
\end{algorithm}

\begin{figure}[ht]
\centering
\vspace{-0.1in}
\includegraphics[width=1.0\linewidth]{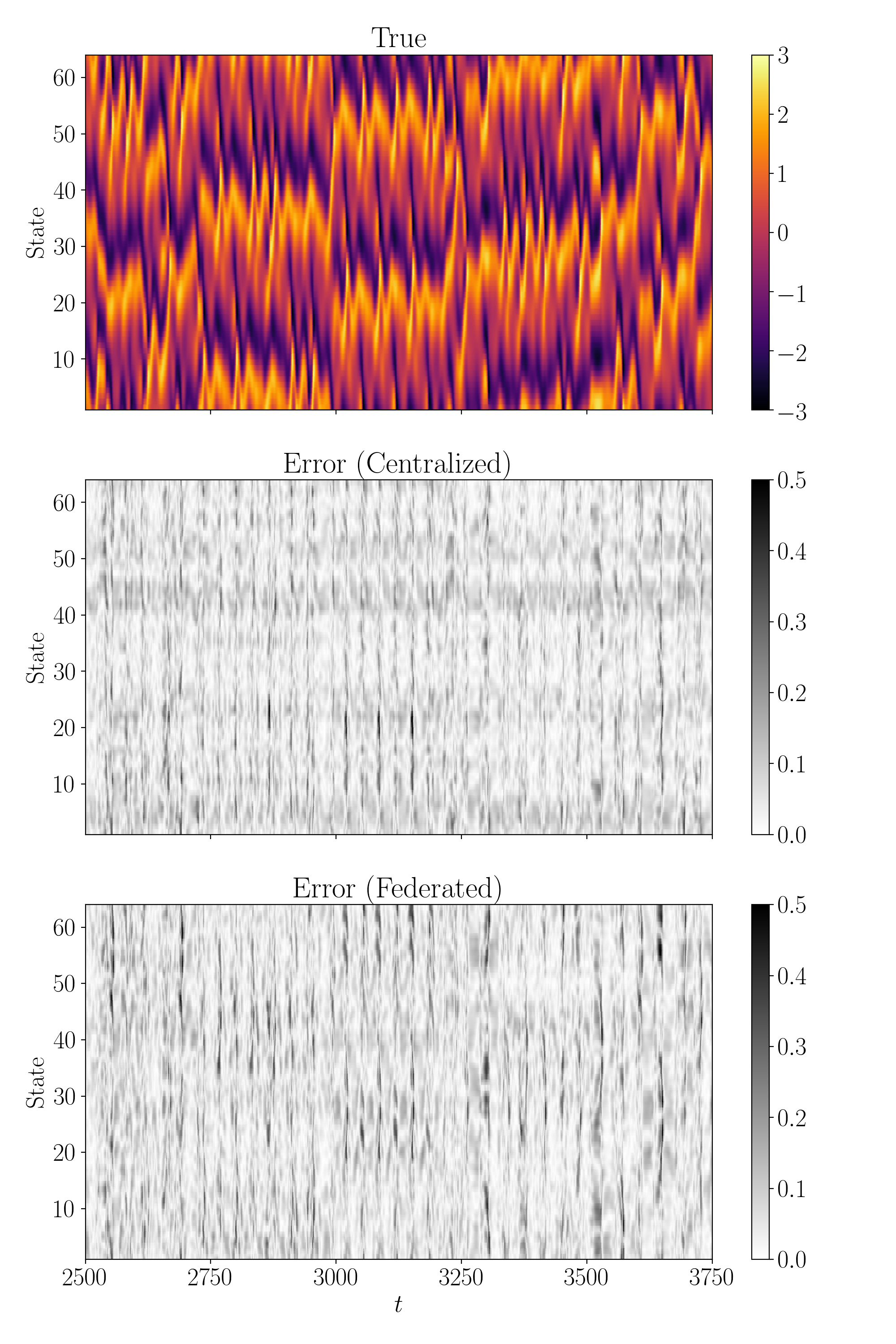}
\vspace{-0.3in}
\caption{Reconstruction performance of the centralized and federated learning approaches with $R=8$.} \vspace{-0.1in}
\label{fig:field}
\end{figure}

The trajectory of the KS system for the testing period is shown in Figure~\ref{fig:field} along with the error between the true data and reconstructed data from centralized and federated autoencoders. The error is computed as the absolute difference between the true and predicted state of the KS system. Both the centralized and federated autoencoders have a similar level of error. 

\emph{Conclusion} --- This letter explores the potential of federated ML for modeling complex spatiotemporal dynamical systems. In particular, we considered the problem of nonlinear dimensionality reduction of chaotic systems as a demonstration case. Federated learning allows for collaborative training of a model while keeping the training data decentralized. Our numerical experiments with the application of autoencoder to the Kuramoto-Sivashinsky system show that a federated model can achieve the same level of accuracy as the model trained using the central data collected from all clients. This work opens up the possibility of updating a model in a centralized setting without exposing the local data collected from different sources. 

\textcolor{rev1}{
We argue that federated learning can solve some of the \emph{big data} challenges in complex dynamical systems provided that the different stakeholders, clients, and vendors use the same \textit{vocabulary} as follows:
\begin{itemize}
\item \textit{Big volume and velocity:} Since inference, analysis and modeling happened on the edge devices only, small amount of data needs to be communicated. This decentralizing process will significantly reduce the communication bandwidth and storage burden. 
\item \textit{Big variety, venue, value and vagueness:} Currently, a lack of trained personnel (to deal with a large variety of data in a centralized location) hinders the adoption of scalable digital solutions. However, the problem is automatically remedied due to domain experts' presence at the data generation venue to extract value, thereby minimizing vagueness.
\item \textit{Big variability, veracity and validity:} The variability in the data generation and sharing processes  resulting from rapid changes in sensor technologies and corresponding regulatory environment will not be a challenge as it will be dealt with locally with federated learning.
\item \textit{Solving data privacy and security issues:} Since the data never leaves the local servers, it will enhance security and encourage clients and vendors to collaborate.  
\end{itemize}
}
Although in this letter we primary focus on federated learning in the context of spatiotemporal prediction of such chaotic systems, our approach can be generalized to large-scale computational settings beyond transport phenomena, for which the research outcomes might improve broader modeling and simulation software capabilities to design cohesive, effective, and secure predictive tools for cross-domain simulations in the various levels of information density. In our future studies, we plan leveraging the decentralized learning approaches in the context of precision meteorology, and develop new physics-guided federated learning approaches to forge new surrogate models compatible among heterogeneous computing environments.

This material is based upon work supported by the U.S. Department of Energy, Office of Science, Office of Advanced Scientific Computing Research under Award Number DE-SC0019290. O.S. gratefully acknowledges their Early Career Research Program support. 

\smallskip
\noindent \textbf{Data Availability}\\
The data that supports the findings of this study is available within the article.

\bibliography{references}

\end{document}